\documentclass[conference]{IEEEtran}
\usepackage[T1]{fontenc}
\usepackage{amsmath,amssymb,booktabs,graphicx,url,array}
\usepackage[hidelinks]{hyperref}
\usepackage{microtype}
\usepackage{algorithm}
\usepackage{algpseudocode}
\usepackage{makecell}
\usepackage[caption=false,font=footnotesize]{subfig}
\graphicspath{{spiketoken_figures/}{figures/}}
\algnewcommand{\Note}[1]{\Statex\hspace{\algorithmicindent}\hspace{\algorithmicindent}$\triangleright$ #1}
\algblockdefx[When]{When}{EndWhen}[1]{\textbf{when} #1}{\textbf{end when}}

\title{SpikeToken: Learning Subword Vocabularies and Tokenizing Bytes with Spiking Networks}

\author{\IEEEauthorblockN{Dalton Diez}
\IEEEauthorblockA{Louisiana State University\\
Baton Rouge, LA, USA\\
ddiez2@lsu.edu}
\and
\IEEEauthorblockN{James Ghawaly}
\IEEEauthorblockA{Louisiana State University\\
Baton Rouge, LA, USA\\
jghawaly@lsu.edu\\
}}

\begin{document}
\maketitle

\setlength{\dbltextfloatsep}{11pt plus 2pt minus 2pt}

\begin{abstract}
Neuromorphic language models have recently garnered attention as a way to improve the efficiency of generative language modeling. These models require tokenization algorithms to convert incoming bytes to tokens; however, most use conventional software tokenizers to perform this task outside of the neuromorphic pipeline. In this work, we introduce SpikeToken, a tokenizer that uses spike dynamics to learn subword vocabularies and segment bytes into tokens. During learning, local synaptic plasticity tracks the frequency of adjacent token pairs. The synaptic strengths determine how pairs are merged into token neurons. During inference, the frozen recurrent network performs spike-based tokenization using these learned synaptic routes. Evaluation on ~1 MB of English–Spanish training text shows that SpikeToken shared 65,502 of 65,536 token spellings (99.95\%) with standard byte-pair encoding (BPE). On all eight vocabulary sizes from 512 to 65,536, SpikeToken reconstructed all 596 held-out documents in our test corpus, byte for byte. We also verified that SpikeToken performs inference tokenization in Lava’s Loihi 2 simulator, the SpiNNaker2 SDK simulator, and on a physical Teensy 4.0 board running the Lacuna C runtime. Overall, tokenization is a critical component of language modeling and SpikeToken narrows the gap towards an end-to-end neuromorphic solution. However, we emphasize that current neuromorphic hardware is insufficient for practical-scale language modeling, with or without spiking tokenization. As such, our work is forward-looking, and because SpikeToken scales with vocabulary size rather than model size, it is well positioned for future neuromorphic language models.
\end{abstract}

\section{Introduction}
In spiking language models, like SpikeGPT\cite{gpt} and SpikeLM\cite{lm}, a tokenizer is needed to convert incoming text into input tokens \cite{gpt,lm}. When a software tokenizer performs this step, a separate processor reads the bytes, determines token boundaries, and passes the resulting token IDs to the model that is ideally running on neuromorphic hardware. However, a spike-driven tokenizer would bring that decision onto neuromorphic hardware, allowing the network to receive byte spikes and supply token spikes directly to the language model.

Most modern tokenizers are open-vocabulary subword tokenizers. The most famous of which is Byte-pair encoding (BPE), which builds subwords by repeatedly merging the most frequent pair of adjacent tokens in the training corpus \cite{bpe,wang}. Each merge changes which tokens are adjacent and therefore which pair may be selected next. Performing such a task in a spiking network poses some challenges; notably, evidence for a pair arrives one occurrence at a time but the merge decision depends on its frequency across the corpus. In our work, we ask whether local synaptic updates can accumulate that evidence and a competition among spiking neurons can select merges. We also ask if a frozen network of a learned vocabulary can perform token segmentation using spike dynamics. 

We use Lacuna to develop SpikeToken (for both vocabulary learning and tokenization), an event-driven spiking neural network simulator with a portable C runtime~\cite{lacuna}. During training, Lacuna updates synaptic weights as token spikes arrive and simulates the competition that selects pairs to merge. After training, it runs the frozen network one byte spike at a time. The network accumulates evidence, then emits a spike pattern that identifies the longest matched token.

Our contribution is the spike-driven learning and token decision rule leveraging the Lacuna SNN simulator \cite{lacuna}. We evaluate it in three stages. First, we compare vocabulary learning across merge per sleep counts, measuring training time and vocabulary overlap with BPE as the reference. Second, we freeze the graph and compare its segmentation with BPE, WordPiece, and Unigram, including cases where nearly identical vocabularies yield different token boundaries. Third, we evaluate the frozen Lacuna network in the Lava Loihi 2 and SpiNNaker2 simulators and run it on a Teensy 4.0 using Lacuna C to demonstrate compatibility with neuromorphic systems and deployment on accessible edge devices. 

We also create an optimized deployment kernel for the CPU and the Teensy 4.0 that evaluates the same gates with bit-parallel operations, preserving token IDs and boundaries on the tested inputs. Finally, the learned routes also support a trie export for fast software-based tokenization.

% Requires \usepackage{graphicx} and \usepackage{subcaption} in the preamble.
% The images are drawn at final print size (8 pt text), so they are included
% without width= scaling. Edit the path prefix to match where the files live.
\begin{figure*}[t]
  \centering
  \subfloat[Round 1\label{fig:lc-round1}]{\includegraphics[width=.4\textwidth,trim={0 5 0 5},clip]{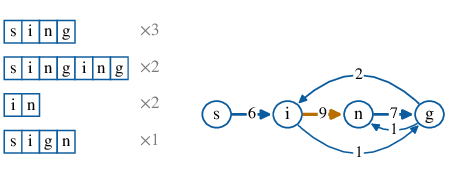}}\hfil
  \subfloat[Round 2\label{fig:lc-round2}]{\includegraphics[width=.4\textwidth,trim={0 5 0 5},clip]{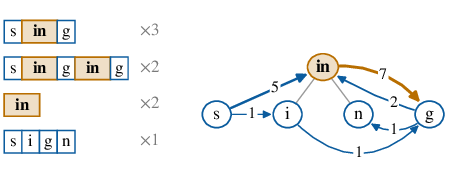}}\\[2pt]
  \subfloat[Round 3\label{fig:lc-round3}]{\includegraphics[width=.4\textwidth,trim={0 5 0 0},clip]{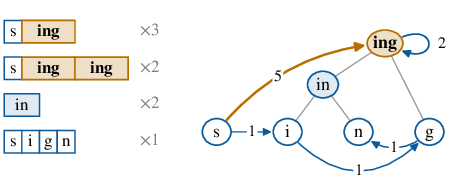}}\hfil
  \subfloat[Round 4\label{fig:lc-round4}]{\includegraphics[width=.4\textwidth,trim={0 5 0 0},clip]{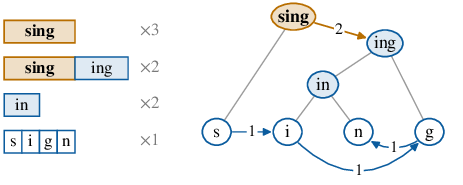}}\\[2pt]
  \subfloat[Final network\label{fig:lc-network}]{\includegraphics[width=.2\textwidth,trim={0 2 0 0},clip]{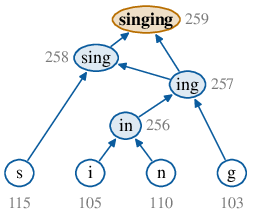}}\hfil
  \subfloat[Vocabulary\label{fig:lc-vocab}]{\includegraphics[width=.2\textwidth,trim={0 2 0 0},clip]{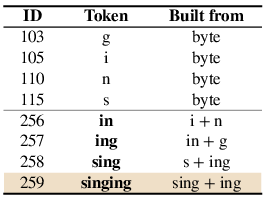}}\hfil
  \subfloat[Final tokenization\label{fig:lc-reading}]{\includegraphics[width=.2\textwidth,trim={0 2 0 0},clip]{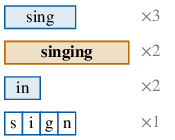}}\\[4pt]
  \includegraphics{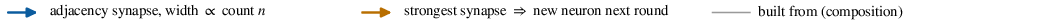}
  \caption{Learning cycles with $k{=}1$ on a toy corpus of eight words.
    (a)--(d)~In each round, the left side shows the corpus as the network reads it, as a sequence of token spikes, and the right side shows the counts on the synapses after reading. Blue arrows are adjacency synapses, with width proportional to count $n$. The orange arrow is the strongest synapse, which becomes a new neuron (orange) in the next round. Gray lines show composition. (e)~Final network: each learned neuron is driven by the two parts it was built from. Gray numbers are token IDs. (f)~Vocabulary: byte IDs 0--255 (four shown) and learned IDs from 256. (g)~With the final vocabulary, the corpus takes 11 token spikes instead of 34. \textit{sign} stays as bytes because none of its pairs was ever the strongest synapse.}
  \label{fig:learning-cycles}
\end{figure*}

\section{Related Work}
\paragraph{Subword Tokenization} The ubiquitous Byte-pair encoding (BPE) tokenizer builds a vocabulary by repeatedly merging the most frequent pair of adjacent symbols. It began as a compression method \cite{gage} and was later adapted for machine translation \cite{bpe}. Other subword tokenizers include WordPiece, whose merge rules maximize co-occurrence likelihood relative to the product of the individual token probabilities rather than pair frequency~\cite{wp,fastwp}. Whereas vanilla BPE and WordPiece are deterministic, others can be stochastic. An example is Unigram, which generates a large vocabulary of all substrings in a corpus and prunes it when doing so maximizes the corpus' likelihood~\cite{uni}. Despite many subword tokenizers being available, BPE (and its variants) remain the most common choice for language model ingestion, for example, as in the case of OpenAI's tiktoken~\cite{openai_tiktoken_2022}. SpikeToken starts from bytes, giving it a shared alphabet across writing systems \cite{wang}. During learning, it adapts BPE’s merge rule by encoding adjacent-pair frequency in synaptic weights. During inference, the frozen SpikeToken network uses spike competition to implement WordPiece’s longest-match rule. Unigram is evaluated as a reference.

\paragraph{Spiking Language Models and Tokenization} SpikeGPT uses a spiking network for next token prediction; however, it does not use a spiking tokenizer. Its smaller model receives character-level inputs while its larger model receives subword IDs from a BPE tokenizer \cite{gpt}. SpikeLM uses spiking networks for language understanding and generation~\cite{lm}. In its BERT-based experiments, a WordPiece tokenizer segments the text before processing it with the spiking language model\cite{lm}. Finally, SPEAK~\cite{speak} uses a leaky integrate-and-fire neuron atop a non-spiking character encoder to predict token boundaries, trained jointly with a downstream language model to reduce its task loss. As such, It requires a language model during training and, for pretrained models, a hypernetwork to embed its open-ended segments. In contrast, SpikeToken learns a fixed subword vocabulary from corpus statistics through local plasticity, uses spike dynamics for both vocabulary learning and segmentation, and runs on neuromorphic simulators and edge hardware.

% \paragraph{Neural Mechanisms} In spike-timing-dependent plasticity (STDP), the relative timing of two neurons’ spikes determines how the synapse between them changes \cite{stdp}. Competition among spiking neurons can also lead different neurons to specialize in different input features \cite{diehl}. SpikeToken's learning methodology for token pair merging is inspired by these ideas, or more broadly, Hebbian principles.

\section{Vocabulary Learning}
\label{sec:train}
SpikeToken performs vocabulary learning in two phases. The first is a ``wake'' phase where the network gathers evidence for the frequency of adjacent token pairs and represents that evidence as synaptic strength. This evidence is gathered during a full pass of the training corpus ingested as bytes. The second phase, ``sleep'', begins after the full pass. In this stage the network is cleared of all spike activity and weights are frozen. Each pair observed during wake is assigned a \emph{candidate} neuron, a temporary neuron that represents the token the pair would form if merged. A single dedicated probe neuron raises each candidate's membrane voltage in proportion to that pair's learned synaptic strength. The first eligible candidate to pass threshold and fire is merged into a new token. SpikeToken's learning mechanism draws inspiration from Hebbian plasticity and competitive spiking networks
\cite{stdp,diehl}.

There can be up to $k$ merges per sleep. An example of these phases with $k=1$ is seen in Figure~\ref{fig:learning-cycles}. On the toy corpus of eight words, during the first pass the \textit{i}$\to$\textit{n} count reaches nine. Before round 2, probe spikes into the network cause the candidate for \textit{i} and \textit{n} to fire first, creating token \textit{in}. This process is repeated in round 2 to create \textit{ing}, round 3 to create \textit{sing}, and round 4 to create \textit{singing}. The final network is seen in Figure~\ref{fig:lc-network}, which represents the learned vocabulary shown in Figure~\ref{fig:lc-vocab}. The final segmentation after vocabulary learning shrinks 34 byte spikes into 11 (Figure~\ref{fig:lc-reading}). 
% We discuss vocabulary learning in more detail and summarize it in Algorithm~\ref{alg:wake-sleep} below.

\begin{algorithm}[htbp]
\caption{SpikeToken vocabulary learning (wake--sleep)}
\label{alg:wake-sleep}
\small
\begin{algorithmic}[1]
\Require Training bytes $S$, target size $V$, merges per sleep $k$
\Ensure Vocabulary $\mathcal{V}$ with $|\mathcal{V}| = V$
\Procedure{Learn}{$S, V, k$}
  \State Create 256 byte neurons; $\mathcal{V} \gets \{0,\dots,255\}$
  \State Encode $S$ as byte spikes, one per position
  \State $z_{ab} \gets 0$ and $\mathcal{A} \gets$ all pairs \Comment{$\mathcal{A}$: pairs to count}
  \While{$|\mathcal{V}| < V$}
    \State \Call{Wake}{$\mathcal{A}$}
    \State $\mathcal{A} \gets$ \Call{Sleep}{$k$}
  \EndWhile
\EndProcedure
\Statex
\Procedure{Wake}{$\mathcal{A}$} \Comment{plasticity on}
  \For{each adjacent $(a,b) \in \mathcal{A}$} \Comment{Eq.~(2)}
    \State $z_{ab} \gets z_{ab} + \bar{\eta}\,(1 - z_{ab})$ \Comment{Eq.~(3)}
  \EndFor
\EndProcedure
\Statex
\Procedure{Sleep}{$k$} \Comment{plasticity off}
  \State Create candidate neuron $c_{ab}$ for each observed pair
  \Note{Probe-to-$c_{ab}$ weight $w_{ab} = 0.9\,z_{ab}$}
  \State Trigger the probe; record candidate firing order
  \State $n \gets \min(k,\, V - |\mathcal{V}|)$
  \State $P \gets$ first $n$ eligible candidates to fire
  \Note{Eligible: $p_a \Vert p_b \notin \mathcal{V}$ and at most 31 bytes}
  \For{each $(a,b) \in P$, in firing order}
    \State Create parent neuron $t$ with spelling $p_a \Vert p_b$
    \State Add $t$ to $\mathcal{V}$
    \State Wire child neurons $a, b$ to $t$ (coincidence circuit)
    \State Replace adjacent $(a,b)$ spikes with one $t$ spike
  \EndFor
  \State $\mathcal{A} \gets$ pairs containing a merged child or new $t$
  \State $z_{ab} \gets 0$ for all $(a,b) \in \mathcal{A}$
  \State \Return $\mathcal{A}$
\EndProcedure
\end{algorithmic}
\end{algorithm}

\subsection{Wake and Sleep}

The initial vocabulary contains one token neuron for each of the 256 possible byte values. New tokens are created by merging ordered pairs of existing tokens. Each learned token $t=(a,b)$ receives a unique ID and represents the concatenated byte spelling $p_t=p_a\Vert p_b$. To be eligible for merging, the resulting spelling must not already exist in the vocabulary and it must not exceed 31 bytes. Algorithm~\ref{alg:wake-sleep} summarizes the vocabulary learning process.

\paragraph{Wake} The objective of the wake phase is to encode the occurrence frequency of each ordered pair of adjacent tokens $(a,b)$ into its corresponding synaptic weight. During the first wake phase, the corpus is represented as a spike stream with one spike per byte. In subsequent wake phases, this representation includes the tokens learned from preceding sleep phases.

Token spikes are timed based on their starting byte positions within their corresponding spellings. As an example, if token $a$ spans three bytes and begins at byte $i$, token $b$ must begin at byte $i+3$ for the two tokens to be adjacent. This temporal representation allows the network to detect adjacency through synaptic delays and decaying membrane voltages.

Each token has an associated ``context'' neuron whose membrane voltage provides a temporal reference for its most recent spike. A token spike resets its context neuron's voltage to $u_r=0.5$, which then decays toward $u_0=0$ with a time constant of $\tau_c=64$ ms. After a duration of time $\Delta$, the context voltage is given by

\begin{equation}
    u_c(\Delta)=u_0+(u_r-u_0)e^{-\Delta/\tau_c}.
\end{equation}

For each candidate pair $(a,b)$, the token with the higher ID sends a query that samples the other token's context voltage. The query delay is configured such that adjacent occurrences produce an expected voltage $u^*_{ab}$. Synaptic plasticity is then gated by an adjacency-dependent eligibility term,

\begin{equation}
    E=\max\!\left(0,1-\max\!\left(0,
    \frac{|u_c-u^*_{ab}|}{\epsilon}-1\right)\right),
\end{equation}

where $\epsilon=10^{-5}$. With the chosen spike timing, $E=1$ for adjacency and $E=0$ otherwise. 
To compute pair frequency, each eligible event increases the corresponding synaptic weight by a fixed fraction of its remaining distance to an upper bound. We parameterize the weight as $w=0.9z$. Where $N_{\mathrm{bytes}}$ is the number of training bytes, we define the effective learning rate as $\bar{\eta}=e^{-1}\min(0.003,1/N_{\mathrm{bytes}})$. Beginning from $z_0=0$, the synaptic update and its closed-form solution after $n$ adjacent occurrences are given as

\begin{equation}
    z^+=z+\bar\eta E(1-z), \qquad z_n=1-(1-\bar\eta)^n.
\end{equation}

As such, synaptic strength is a monotonically increasing function of adjacent-pair frequency. This allows the network to rank candidate merges through neuronal competition, confined to numerical precision limit.

\paragraph{Sleep} The objective of the sleep phase is to identify the strongest adjacency synapses and incorporate their corresponding token pairs into the vocabulary. In this step, plasticity is disabled, and a temporary candidate neuron is created for each observed pair. A shared probe neuron is wired to each candidate with a weight set to the learned adjacency synapse.

A trigger spike starts a competition among the candidate neurons. Candidates with stronger incoming synapses reach their firing thresholds more quickly under the same drive. As such, the resulting firing order ranks candidate pairs by their learned synaptic strengths. Ties are resolved by the simulator's event ordering. Any pair whose combined spelling already exists or exceeds 31 bytes is skipped. This process is continued until the vocabulary target $V$ is reached.

One parent token neuron $t$ is created for each accepted pair $(a,b)$ by the host, wiring its two child neurons through a coincidence circuit. This circuit detects adjacent occurrences of $a$ and $b$ in the current spike stream, suppresses their individual spikes, and generates a single parent spike. Inhibition prevents overlapping merges from absorbing the same spike. This could occur, for example, in the sequence $(a,a,a)$.

These merges change the adjacency relationships within the corpus. Before starting the next wake phase, synaptic weights associated with affected pairs are reset so that their frequencies can be recomputed. The updated spike stream then becomes the input to the next wake phase.

\paragraph{Sleep Merge Frequency} The parameter $k$ controls the trade-off between merge selection fidelity to BPE and training time. When $k=1$, pair frequencies are updated after every merge, as is done in BPE \cite{bpe}. When $k>1$, multiple merges are selected from the same wake-phase statistics, reducing the number of wake–sleep cycles required to reach the target vocabulary size. This results in faster training because it reduces the number of cycles needed to reach $V$. However, an earlier merge may alter the frequency of a pair selected for a subsequent merge within the same sleep phase.

\subsection{Training and Baselines}\label{sec:training}

We trained SpikeToken on 8,098 documents containing 1,046,545 bytes of aligned English--Spanish OPUS Books text \cite{opus}. We split the aligned text by bytes before training and gave all four tokenizers the same raw bytes without normalization or pre-tokenization. For each SpikeToken merge schedule ($k=1,16,64$), we recorded cumulative training time as $V$ grew. At each checkpoint, we compared the learned byte spellings with those of BPE, WordPiece, and Unigram, each trained independently at the same requested vocabulary size using Hugging Face Tokenizers 0.23.2.

\begin{figure}[htpb]
\centering
\includegraphics[width=.8\columnwidth]{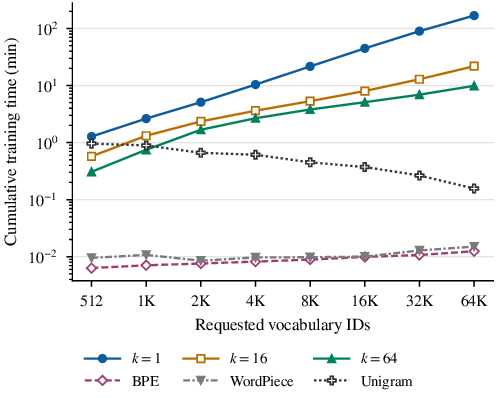}
\caption{Cumulative wall time to reach each measured vocabulary size. SpikeToken timing includes simulation, routing, and graph construction.}
\label{fig:training}
\end{figure}

Figure~\ref{fig:training} shows the training time on a single core of an AMD Ryzen~9 7950X3D CPU at each requested vocabulary size. At $V=32768$, SpikeToken took 89.68 minutes with $k=1$, 12.9 minutes with $k=16$, and 6.9 minutes with $k=64$. Relative to \(k=1\) these times show speedups of $6.95\times$ for \(k=16\) and $12.97\times$ \(k=64\) . At the same vocabulary size, median training times over five fits were 1.39 seconds for byte BPE, 1.44 seconds for WordPiece, and 34.72 seconds for Unigram. 
% We note two things: first, training need only be performed once prior to deployment and second, the Hugging Face Tokenizers implementations of BPE, WordPiece, and Unigram represent years of software engineering and optimization. Similar efforts could be applied to SpikeToken to optimize training speed but that is not the objective of this work.

Table~\ref{tab:overlap} reports the percentage of SpikeToken spellings shared with each software tokenizer, including the 256 initial byte tokens With $k=1$, SpikeToken matched every BPE spelling through $V=8192$ and 65,502 of 65,536 spellings at the largest vocabulary size. Increasing $k$ reduced vocabulary overlap while accelerating training. Overlap with Unigram was much lower, which is expected due to differences in its learning algorithm.

% Table~\ref{tab:overlap} reports the percentage of SpikeToken spellings shared with each software tokenizer, including the 256 initial byte tokens. We note that each of these tokenizers uses a unique algorithm. As such, the intention of this comparison is not to maximize shared spellings but to understand how their alignment changes as $V$ and $k$ are adjusted. The expectation is that SpikeToken vocabulary aligns the closest with BPE due to SpikeToken's BPE-inspired design. This is, in fact, what we see. SpikeToken with a single merge per sleep cycle ($k=1$) matched every BPE spelling through $V=8192$ and shared 65,502 of 65,536 spellings at the largest vocabulary size. While $k=64$ can significantly reduce training time (see Figure~\ref{fig:training}), it reduces overlap with the baselines. At $V=32768$, overlap with BPE decreases from 99.98\% to 48.86\%, and with WordPiece it decreases from 99.98\% to 48.85\%. Unigram's overlap is low for all but the smallest of vocabulary sizes, which is expected as its algorithm follows a very different methodology. 

\begin{table}[htbp]
\centering
\caption{Percentage of SpikeToken spellings shared with each tokenizer.}
\label{tab:overlap}
\footnotesize
\renewcommand{\arraystretch}{0.9}
\setlength{\aboverulesep}{1pt}
\setlength{\belowrulesep}{1pt}
\setlength{\tabcolsep}{1pt}
\begin{tabular*}{\columnwidth}{@{\extracolsep{\fill}}r*{9}{r}@{}}
\toprule
& \multicolumn{3}{c}{BPE} & \multicolumn{3}{c}{WordPiece} & \multicolumn{3}{c}{Unigram} \\
\cmidrule(lr){2-4}\cmidrule(lr){5-7}\cmidrule(lr){8-10}
IDs & k=1 & k=16 & k=64 & k=1 & k=16 & k=64 & k=1 & k=16 & k=64 \\
\midrule
512    & 100.00 & 90.23 & 81.64 & 99.80 & 90.23 & 81.64 & 78.52 & 78.12 & 75.78 \\
1,024  & 100.00 & 82.03 & 70.21 & 99.90 & 82.03 & 70.21 & 59.38 & 58.98 & 58.30 \\
2,048  & 100.00 & 77.98 & 64.26 & 99.95 & 77.93 & 64.21 & 46.48 & 46.09 & 45.26 \\
4,096  & 100.00 & 74.44 & 61.08 & 99.98 & 74.44 & 61.08 & 36.47 & 38.28 & 37.23 \\
8,192  & 100.00 & 72.39 & 57.73 & 99.99 & 72.39 & 57.73 & 29.28 & 30.92 & 30.29 \\
16,384 &  99.98 & 69.60 & 54.21 & 99.98 & 69.60 & 54.21 & 23.00 & 24.26 & 23.72 \\
32,768 &  99.98 & 65.12 & 48.86 & 99.98 & 65.12 & 48.85 & 23.97 & 23.61 & 22.34 \\
65,536 & 99.95 & 56.86 & 40.25 & 99.93 & 56.87 & 40.25 & 22.78 & 24.04 & 24.16 \\
\bottomrule
\end{tabular*}
\end{table}

\section{Tokenization}
\label{sec:inference}

\begin{figure*}[htbp]
  \centering
  \subfloat[Route circuit of the learned vocabulary\label{fig:circuit-route}]{\includegraphics[width=.8\textwidth,trim={0 2 0 0},clip]{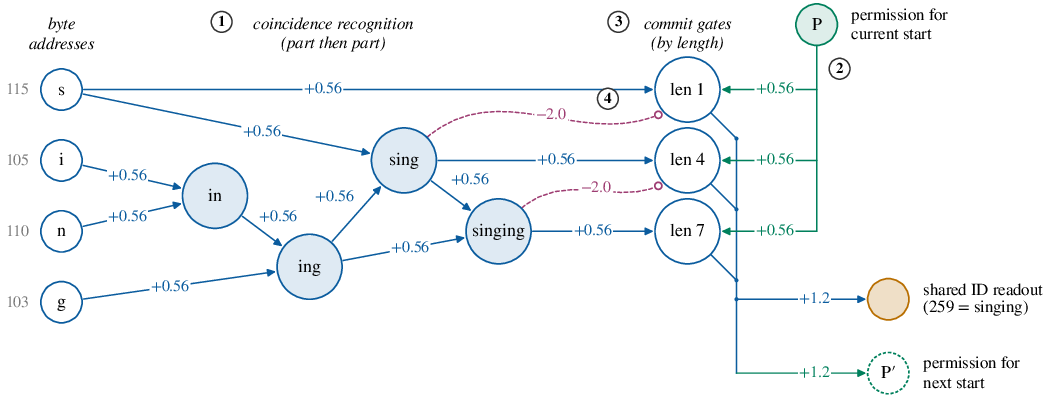}}\\[4pt]
  \subfloat[Tokenizing the input ``singing'' (event order; timing schematic)\label{fig:circuit-timing}]{\includegraphics[width=.8\textwidth,trim={0 2 0 2},clip]{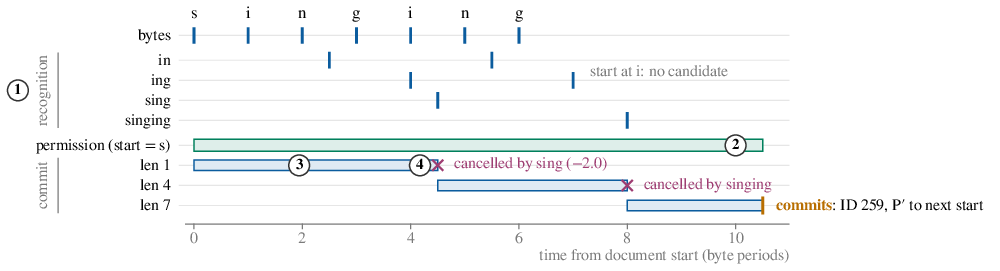}}
  \caption{Frozen Lacuna circuit for \textit{singing}. Solid blue and dashed purple lines are
  excitatory and inhibitory synapses, respectively. 
  % the shared ID readout and to $P'$. Dashed purple lines ending in an open circle inhibit shorter matches at the same start. Green lines carry start permission. 
  $P'$ marks permission for the first byte after the selected token. 
  In~(b), the length-7 gate commits ID~259 at 10.5 byte periods.}
  \label{fig:circuit}
\end{figure*}

After training, the network is frozen to preserve the vocabulary and tokenization can begin, as outlined in Algorithm~2. Each token keeps its learned ID, and the routes of the two children that formed it. 
% These child to parent routes are used during tokenization. For example, in Figure~\ref{fig:learning-cycles}, we saw that the token \textit{in} was learned from \textit{i} and \textit{n}. 
During tokenization, these routes are followed to recognize longer tokens and emits the longest match that begins at the first unread byte.

Figure~\ref{fig:circuit} shows how the vocabulary learned in Figure~\ref{fig:learning-cycles} is used to tokenize \textit{singing}. SpikeToken reads the word from left to right, with one spike per input byte. The first byte is \textit{s}, which is itself a token. A ``permission'' spike marks this position as the start of the first output token. The longest token in this example vocabulary is seven bytes long, so up to seven bytes can affect which token is emitted.

Next we see \textit{i} and \textit{n}. The connection from \textit{i} delays its spike so that the two spikes reach the \textit{in} neuron together and make it fire. We then see \textit{g}, whose spike combines with \textit{in} to recognize \textit{ing}. The \textit{ing} spike combines with the delayed spike from the initial \textit{s} to recognize the four-byte token \textit{sing}. The final three bytes produce a second \textit{ing} spike. This combines with the delayed \textit{sing} spike to recognize the seven-byte token \textit{singing}.

The network aligns the matches that begin at the initial \textit{s} so they can be compared at the same time. At this permitted start, \textit{singing} inhibits the shorter matches \textit{s} and \textit{sing}. The network then emits \textit{singing} (ID~$259$) and advances by seven bytes. We note that permission is propagated by the network; only the first permission is initiated by the host. 
% This process is examined further in Algorithm~\ref{alg:frozen-tokenization}.

\begin{algorithm}[htbp]
\caption{Frozen spike-driven tokenization}
\label{alg:frozen-tokenization}
\small
\begin{algorithmic}[1]
\Require Bytes $x_1,\dots,x_N$; frozen network; longest spelling $L$
\Ensure Token IDs by greedy longest match
\Statex \textit{Host}
\State Stream byte spikes $x_1,\dots,x_N$, one per byte period
\State Fire permission neuron $P$ at byte 1
\State Record readout spikes until the network is quiet
\Statex
\Statex \textit{Network (frozen; runs on spike events)}
\When{child neurons $a, b$ of token $t$ fire adjacently}
  \State Delayed spikes coincide; recognition neuron $t$ fires
\EndWhen
\When{a recognition neuron of length $\ell$ fires}
  \State Its spike is delayed to the end of its start's window
  \State It excites length gate $g_\ell$
  \State An inhibitory neuron suppresses $g_{\ell'}$, $\ell' < \ell$
\EndWhen
\When{gate $g_\ell$ has a match and permission at byte $i$}
  \State $g_\ell$ fires; only the longest match survives
  \State Readout neuron emits the winning token's index
  \State $P'$ fires and grants permission at byte $i + \ell$
\EndWhen
\end{algorithmic}
\end{algorithm}

Algorithm~\ref{alg:frozen-tokenization} begins with a permission spike at the first byte of the document (\textit{s} in \textit{singing}). The permission spike identifies where a new token must begin. We then know from training the longest token byte length available from that beginning token. We refer to this as window size and we search inside this window for the longest token match. 
% SpikeToken then reads the input from left to right, sending one spike per byte to its corresponding neuron.

For a learned token $t=(a,b)$, synaptic delays align the two child spikes at the parent neuron when $b$ immediately follows $a$. Each spike alone is below threshold, but together they make the parent fire. 
% For example, the connection from \textit{i} delays its spike one byte interval longer than the connection from \textit{n}, so adjacent \textit{i} and \textit{n} activate \textit{in}.

Matches that begin at the same byte may be recognized at different times. Fixed delays align them for comparison. 
% Each recognized length inhibits the gates for shorter lengths at that start. In the example, \textit{singing} inhibits both \textit{sing} and \textit{s}, leaving only the seven-byte match eligible for output. Fixed synaptic delays align matches that begin at the same byte. 
Each match excites its length gate with weight $+1.2$ and, through an inhibitory neuron, suppresses shorter-length gates with weight $-2.0$. This keeps the shorter gates below the firing threshold of $1.0$. At the end of the window, the surviving match and the start permission together bring the commit neuron above threshold, causing the network to emit the token and pass permission to the first byte after it, repeating the cycle. If no learned token matches at the permitted start, the network emits the corresponding byte token from the initial vocabulary of 256 bytes and advances permission by one byte.

\begin{table}[t!]\centering
\caption{Metrics at $V=32768$ IDs on all 596 evaluation documents. Entropies use the ID distribution, normalized by $\log_2 V$.}\label{tab:quality}
\footnotesize\setlength{\tabcolsep}{2.8pt}
\renewcommand{\arraystretch}{0.9}
\setlength{\aboverulesep}{1pt}
\setlength{\belowrulesep}{1pt}
\begin{tabular}{@{}lrrrrrr@{}}\toprule
\makecell[lb]{Tokenizer} &
\makecell[rb]{Token\\Count} &
\makecell[rb]{Token Byte\\Density} &
\makecell[rb]{Vocab.\ Use\\(\% IDs Used)} &
\makecell[rb]{Fertility} &
\makecell[rb]{$H_1$} &
\makecell[rb]{$H_{2.5}$}\\\midrule
\makecell[l]{SpikeToken\\($k=1$)}  & 93,630  & 2.815 & 31.81 & 1.646 & 0.631 & 0.278 \\
\makecell[l]{SpikeToken\\($k=16$)} & 93,018 & 2.834 & 32.50 & 1.660 & 0.642 & 0.285 \\
\makecell[l]{SpikeToken\\($k=64$)} & 92,857  & 2.839 & 34.02 & 1.722 & 0.660 & 0.308 \\
BPE         & 94,395  & 2.793 & 30.12 & 1.693 & 0.631 & 0.288 \\
WordPiece   & 93,688  & 2.814 & 31.81 & 1.646 & 0.631 & 0.278 \\
Unigram     & 102,607 & 2.569 & 25.41 & 1.736 & 0.590 & 0.287 \\
\bottomrule\end{tabular}\end{table}

\subsection{Experiments}

Here, We evaluate SpikeToken's byte-level reconstruction fidelity and compare its segmentation characteristics against conventional tokenizers. 
% We also qualitatively analyze a selection of tokenizations and compare them against the reference methods.

\paragraph{Data} We evaluated segmentation on 538 held-out English and Spanish passages from OPUS Books~\cite{opus}. To test how the learned vocabulary generalizes beyond literary prose, we included 30 samples from the CPython 3.13.0 source code and the AnCora Spanish news corpus~\cite{ancora}. A further 28 manually crafted byte-level stress cases, including empty input, malformed UTF-8, and a sequence containing all 256 byte values, tested whether tokenization preserved the input exactly.

\paragraph{Metrics}
We ran the trained tokenizers on all 596 evaluation documents. Every evaluated SpikeToken checkpoint reconstructed the input byte for byte, including all 28 stress cases. To compare how the tokenizers segmented the same text, we measured document token sequence length, tokenizer word-split frequency, fertility~\cite{rust}, vocabulary use, Shannon entropy, and R\'enyi entropy~\cite{entropy,schmidt}. The purpose of these metrics is outlined as follows:

\begin{itemize}
    \item \textbf{token sequence length}: measures byte-to-token compression. Increased bytes per token decreases token sequence length; however, higher is not always better for downstream learning tasks.
    \item \textbf{token byte density}: measures the average number of bytes per token in generated sequences.
    \item \textbf{fertility}: measures the average number of tokens generated per word for a given text~\cite{rust}.
    \item \textbf{vocabulary use}: measures the fraction of a tokenizer's vocabulary that is used for a given text.
    \item \textbf{Shannon entropy ($H_{1}$)}: measures the token sequence entropy over a given text. Specifically, it provides a measure of vocabulary utilization over a text, describing how a tokenizer's vocabulary is distributed. Higher entropy indicates more uniform vocabulary distribution whereas lower entropy measures higher reliance on a smaller subset of the vocabulary. Higher entropy is generally preferred.
    \item \textbf{R\'enyi entropy ($H_{2.5}$)}: provides a similar measure to Shannon entropy. However, with an order of 2.5 it gives more weight to the most frequent IDs and thus aids in the analysis of vocabulary distribution utilization~\cite{entropy,schmidt}.
\end{itemize}

These metrics were computed on the 596 evaluation documents and are reported in Table~\ref{tab:quality}. 

\paragraph{Analysis}
SpikeToken passed all 28 stress tests. The frozen network retains the original input tokens for every 256 possible byte values. When no longer learned token matches at the current position, the network emits the single-byte token and advances one byte. This preserves malformed UTF-8 exactly. 
% Empty input produces no tokens. 

As seen in Table~\ref{tab:quality}, SpikeToken, BPE, and WordPiece all generally produce similar token counts, ranging from 92{,}857 at the lowest (SpikeToken with $k=64$) and 94{,}395 at the highest. Unigram, however, produces a fair bit more at 102{,}607. This suggests that at this vocabulary size ($V=32768$), SpikeToken, BPE, and WordPiece may more heavily compress content than Unigram. This analysis is consistent with the other metrics as well. For example, Unigram presents the lowest Shannon entropy (($H_{1}=0.590$) of the group, with the others grouped closely and ranging from 0.631 and 0.660. Notably, SpikeToken with $k=64$ produces the highest Shannon entropy among all tokenizers. The story is a bit more subtle for ($H_{2.5}$) however, where SpikeToken ($k=1$), BPE, Wordpiece, and Unigram are clustered together between 0.278 and 0.288 and SpikeToken ($k=64$) produces the highest $H_{2.5}$ entropy at 0.308. The results for token byte density and vocabulary usage are also consistent with SpikeToken ($k=64$) having the highest values across both metrics. Finally, for fertility, we find that the methods all cluster fairly close together, ranging from 1.646 (SpikeToken, $k=1$ and WordPiece) to 1.736 (Unigram).
We further discuss the implications of these results in Section~\ref{sec:discussion}.

\begin{figure*}[htbp]
  \centering
  \subfloat[Token density\label{fig:quality-density}]{\includegraphics[width=.4\textwidth,trim={0 5 0 0},clip]{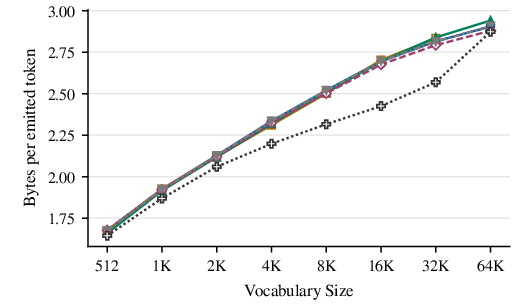}}\hfil
  \subfloat[Fertility\label{fig:quality-fertility}]{\includegraphics[width=.4\textwidth,trim={0 5 0 8},clip]{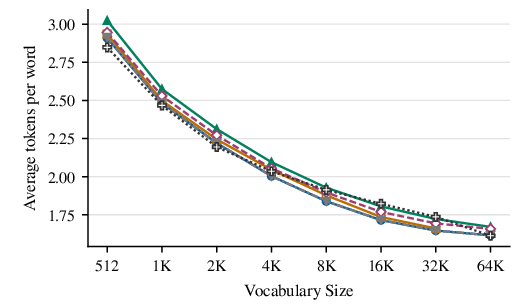}}\\[2pt]
  \includegraphics[width=\textwidth,trim={0 5 0 5},clip]{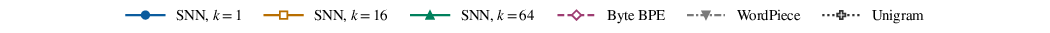}
  \caption{Token byte density and mean fertility for the three SpikeToken SNNs and three software controls on the same 596 documents. Higher density means fewer tokens for the same bytes. Lower fertility means fewer pieces overlap a word.}
  \label{fig:quality}
\end{figure*}

Table~\ref{tab:examples} shows example tokenizations for two passages for SpikeToken ($k=1$ and $k=64$), Wordpiece, Unigram, and BPE. Additionally, Figure~\ref{fig:quality} plots token byte density and fertility against vocabulary size ($V$) for the baselines as well as SpikeToken at three $k$ settings. As expected, we see that token density monotonically increases and fertility monotonically decreases with $V$ for all tokenizers. The only interesting observation is that Unigram reports noticeably lower token density than the others between 1K and 32K vocabulary sizes.

% Requires booktabs. Insert with \input{spiketoken_examples}.
\begin{table*}[t]
\centering
\caption{Token boundaries for two held-out passages at 32,768 vocabulary IDs. A vertical bar separates tokens, \texttt{\textvisiblespace{}} marks a space, and \texttt{\textbackslash{}n} marks the final newline. SpikeToken ($k=1$) and WordPiece share a row because their boundaries agree throughout both passages.}
\label{tab:examples}
\footnotesize
\newcommand{\STboundary}{\,\textbar\,}
\begin{tabular}{@{}lr@{\hspace{1em}}l@{}}
\toprule
Reader & Tokens & Emitted pieces in order \\
\midrule
\multicolumn{3}{@{}l}{\emph{English}\quad\texttt{Presently\textvisiblespace{}he\textvisiblespace{}addressed\textvisiblespace{}me-{}-\textquotedbl{}Your\textvisiblespace{}name,\textvisiblespace{}little\textvisiblespace{}girl?\textquotedbl{}\textbackslash{}n}} \\
SpikeToken ($k=1$), WordPiece & 9 & \texttt{Present\STboundary{}ly\textvisiblespace{}he\textvisiblespace{}\STboundary{}addressed\textvisiblespace{}\STboundary{}me-{}-\STboundary{}\textquotedbl{}Your\textvisiblespace{}\STboundary{}name,\textvisiblespace{}\STboundary{}little\textvisiblespace{}\STboundary{}girl\STboundary{}?\textquotedbl{}\textbackslash{}n} \\
SpikeToken ($k=64$) & 11 & \texttt{Present\STboundary{}ly\textvisiblespace{}\STboundary{}he\textvisiblespace{}\STboundary{}addres\STboundary{}sed\textvisiblespace{}\STboundary{}me-{}-\STboundary{}\textquotedbl{}Your\textvisiblespace{}\STboundary{}name,\textvisiblespace{}\STboundary{}little\textvisiblespace{}\STboundary{}girl\STboundary{}?\textquotedbl{}\textbackslash{}n} \\
BPE & 10 & \texttt{Pres\STboundary{}ently\textvisiblespace{}\STboundary{}he\textvisiblespace{}\STboundary{}addressed\textvisiblespace{}\STboundary{}me-{}-\STboundary{}\textquotedbl{}Your\textvisiblespace{}\STboundary{}name,\textvisiblespace{}\STboundary{}little\textvisiblespace{}\STboundary{}girl\STboundary{}?\textquotedbl{}\textbackslash{}n} \\
Unigram & 10 & \texttt{Presently\STboundary{}\textvisiblespace{}he\textvisiblespace{}\STboundary{}addressed\textvisiblespace{}\STboundary{}me-{}-\textquotedbl{}\STboundary{}Y\STboundary{}our\textvisiblespace{}name\STboundary{},\textvisiblespace{}\STboundary{}little\textvisiblespace{}girl\STboundary{}?\textquotedbl{}\STboundary{}\textbackslash{}n} \\
\midrule
\multicolumn{3}{@{}l}{\emph{Spanish}\quad\texttt{Al\textvisiblespace{}fin,\textvisiblespace{}me\textvisiblespace{}preguntó:\textvisiblespace{}-\textquestiondown{}Cómo\textvisiblespace{}te\textvisiblespace{}llamas,\textvisiblespace{}niña?\textbackslash{}n}} \\
SpikeToken ($k=1$), WordPiece & 11 & \texttt{Al\textvisiblespace{}\STboundary{}fin,\textvisiblespace{}\STboundary{}me\textvisiblespace{}\STboundary{}pregunt\STboundary{}ó:\textvisiblespace{}-\STboundary{}\textquestiondown{}Cómo\textvisiblespace{}\STboundary{}te\textvisiblespace{}\STboundary{}llam\STboundary{}as,\textvisiblespace{}\STboundary{}niñ\STboundary{}a?\textbackslash{}n} \\
SpikeToken ($k=64$) & 11 & \texttt{Al\textvisiblespace{}\STboundary{}fin,\textvisiblespace{}\STboundary{}me\textvisiblespace{}pregunt\STboundary{}ó\STboundary{}:\textvisiblespace{}-\STboundary{}\textquestiondown{}Cómo\textvisiblespace{}\STboundary{}te\textvisiblespace{}l\STboundary{}lam\STboundary{}as,\textvisiblespace{}ni\STboundary{}ñ\STboundary{}a?\textbackslash{}n} \\
BPE & 12 & \texttt{Al\textvisiblespace{}\STboundary{}fin,\textvisiblespace{}\STboundary{}me\textvisiblespace{}\STboundary{}pregunt\STboundary{}ó\STboundary{}:\textvisiblespace{}\STboundary{}-\textquestiondown{}Cómo\textvisiblespace{}\STboundary{}te\textvisiblespace{}\STboundary{}llam\STboundary{}as,\textvisiblespace{}\STboundary{}niñ\STboundary{}a?\textbackslash{}n} \\
Unigram & 12 & \texttt{Al\textvisiblespace{}fin\STboundary{},\textvisiblespace{}me\textvisiblespace{}\STboundary{}pregunt\STboundary{}ó\STboundary{}:\textvisiblespace{}\STboundary{}-\textquestiondown{}Cómo\textvisiblespace{}\STboundary{}te\textvisiblespace{}\STboundary{}llama\STboundary{}s,\textvisiblespace{}\STboundary{}niña\STboundary{}?\STboundary{}\textbackslash{}n} \\
\bottomrule
\end{tabular}
\end{table*}

\section{Deployability}
\label{sec:deploy}

\begin{figure*}[htbp]
  \centering
  \subfloat[Desktop\label{fig:readers-desktop}]{\includegraphics[width=.4\textwidth,trim={0 5 0 8},clip]{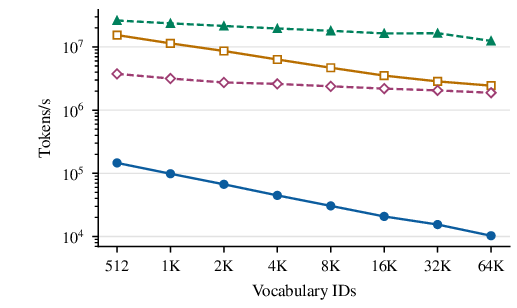}}\hfil
  \subfloat[Physical Teensy 4.0\label{fig:readers-teensy}]{\includegraphics[width=.4\textwidth,trim={0 5 0 8},clip]{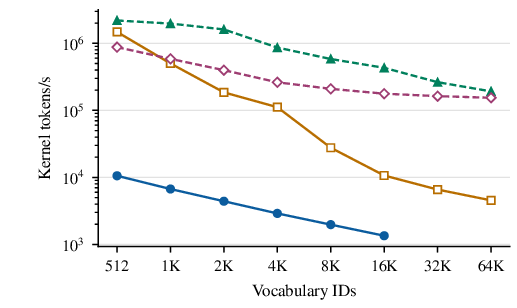}}\\[2pt]
  \includegraphics[width=\textwidth,trim={0 5 0 5},clip]{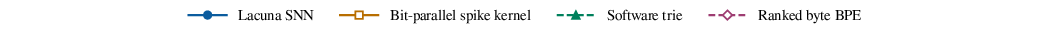}
  \caption{SpikeToken ($k=1$) at seven vocabulary sizes on 596 documents. Left, the base Lacuna network, bit parallel spike SNN, software trie and byte BPE. Right, physical Teensy throughput uses kernel time.}  
  \label{fig:speed}
\end{figure*}

\subsection{Lacuna}
SpikeToken's main implementation is developed in Lacuna~\cite{lacuna}, which simulates it as an event-driven spiking neural network. The host supplies the network with the document-start spike and proceeds with one input spike for each byte. The network then drives the rest of the tokenization process as described in Section~\ref{sec:inference}. The host can either read the winning token spikes for tokenization or those spikes could be provided directly to a downstream neuromorphic language model. On a single desktop core of the machine described in Section~\ref{sec:training}, the Lacuna SpikeToken implementation with $V=32768$ processed the 596-document evaluation set at a rate of 15{,}463 tokens/s. Figure~\ref{fig:speed}a plots this.

Additionally, we used Lacuna's portable C runtime to run SpikeToken directly on a physical Teensy~4.0 mirocontroller (600 MHz ARM Cortex-M7 processor) at vocabulary sizes ranging from $V=512$ to $V=16384$ (32K+ did not fit in RAM or flash storage). Here, throughput ranged from 10{,}570 to 1{,}347 tokens/s, decreasing monotonically with $V$. Figure~\ref{fig:speed}b plots this across the vocabulary sizes evaluated.

\begin{table}[htbp]
\centering
\caption{SpikeToken ($k=1$) circuit size. Lava's neuron and synapse counts
include delay relays and recurrent connections, respectively.
% At every vocabulary size, Lava's fixed-point CPU simulation reproduced
% the native Lacuna token IDs and byte-start positions on the test inputs.
}
\label{tab:devices}
\footnotesize
\renewcommand{\arraystretch}{0.9}
\setlength{\aboverulesep}{1pt}
\setlength{\belowrulesep}{1pt}
\setlength{\tabcolsep}{2pt}
\begin{tabular*}{\columnwidth}{@{\extracolsep{\fill}}r rr rr@{}}
\toprule
& \multicolumn{2}{c}{Lacuna LIF}
& \multicolumn{2}{c}{Fixed-point Lava LIF} \\
\cmidrule(lr){2-3}\cmidrule(lr){4-5}
IDs & Neurons & Synapses & Neurons & Synapses \\
\midrule
512    &    758 &   3,993 &    759 &   3,994 \\
1,024  &  1,361 &   8,290 &  1,368 &   8,297 \\
2,048  &  2,506 &  16,799 &  2,514 &  16,807 \\
4,096  &  4,738 &  33,693 &  4,750 &  33,705 \\
8,192  &  9,034 &  67,106 &  9,052 &  67,124 \\
16,384 & 17,477 & 133,464 & 17,503 & 133,490 \\
32,768 & 34,275 & 265,759 & 34,451 & 265,935 \\
65,536 & 67,401 & 528,707 & 67,589 & 528,895 \\
\bottomrule
\end{tabular*}
\end{table}

Table~\ref{tab:devices} shows how the frozen SpikeToken circuit grows with vocabulary size. Each token needs neurons for recognition and output, and the network also uses neurons to inhibit shorter matches and advance the next allowed start. In the fixed-point Lava circuit, chains of relay neurons supply the delays that make two child spikes reach their parent together. In Lava simulation a connection can delay a spike up to 64 ticks. When a route needs a longer delay, the network must forward it to a different delay neuron. 
% Routes that need parts of the same delay can share neurons.

We checked the same token decisions in Lava's Loihi~2 fixed-point CPU simulation and in the SpiNNaker2 SDK's Brian2 simulator \cite{lava,spinnaker}. Both reproduced every token ID and byte start on the 596 test inputs. The SpiNNaker2 mapper assigned the 512-ID circuit to 6 cores and the 1,024-ID circuit to 9 cores.

\subsection{Conventional CPU Optimizations}
While SpikeToken operates as a spiking neural network targeting neuromorphic hardware, we recognize that users may not have access to such hardware or may wish to run SpikeToken on standard CPU. To support this, we developed two optimized implementations that preserve SpikeToken's tokenization behavior: a bit-parallel spike kernel and a software trie. 
% In this case, various optimizations can be made to substantially increase tokenization rates without impacting operation. We perform two such optimizations, one using a custom spike kernel and the other using a software trie. 
We emphasize that neither of these implementations serve our primary neuromorphic objectives but are provided to improve utility on non-neuromorphic platforms.

\paragraph{Bit-parallel spike kernel}
Once training in Lacuna is complete, the tokenizer can run in a custom bit-parallel kernel. The kernel represents each neuron's firing at up to 32 consecutive byte positions as bits in one machine word. For each learned pair, it shifts the right child's bits by the left child's length and intersects the two words. A surviving bit marks a position where the children are adjacent. Further bit operations suppress shorter matches and allow a token to start only where the previous one ended. This preserves the token decision while replacing individual membrane updates and spike deliveries with word operations.

Across all eight single-merge vocabulary sizes, the kernel matched every token ID and byte start emitted by the explicit network on the 596 desktop inputs. At $V=65536$ IDs, desktop throughput rose from 10{,}279 tokens/s in Lacuna to  2{,}440{,}617 tokens/s with the kernel (Fig.~\ref{fig:speed}, left). On the physical Teensy~4.0, the kernel processed 4,544 tokens/s at that size on all 596 documents. In a separate Lava CPU simulation, custom code evaluated the packed gates and stock LIF neurons emitted the token spikes (a hybrid neuromorphic-CPU implementation). 

\paragraph{Software trie from learned routes}
Each spike-driven merge records the two child tokens that form a new token. Following those links back to the original byte neurons recovers the token's exact spelling. We can therefore export the learned vocabulary as a software trie~\cite{tries}. Tokens with a common byte prefix share a path in the trie, and each complete spelling carries the token ID assigned during training.

% The trie reads from the first unread byte, following one edge per matching byte and remembering the last complete token it reached. When the path ends or the input runs out, it emits that token's ID and resumes immediately after its spelling. The export preserves the frozen SpkeToken network's longest-match rule and requires no further training. During inference, byte comparisons and memory lookups perform the entire segmentation without spikes.

At $V=65536$ IDs, the trie matched the SpikeToken SNN on all 596 documents, emitting the same 90,755 tokens at the same byte positions. On the desktop, the trie processed 12.48 million tokens/s compared with 1.88 million for BPE. On the physical Teensy~4.0, the trie reached 191,892 tokens/s compared with 154,200 for BPE (Fig.~\ref{fig:speed}).

\section{Discussion and Limitations}\label{sec:discussion}

While neuromorphic implementations of language models (particularly LLMs) has recently gained significant attention, to our knowledge, all current implementations rely on conventional software-based tokenization. To enable future end-to-end neuromorphic implementations of LLMs, spiking tokenizers are needed to enable neuromorphic processors to directly ingest text bytes. Our solution to this, SpikeToken, is the first such implementation of a gradient-free spiking tokenizer designed to operate in a similar to that of the ubiquitous BPE algorithm.

Beyond this, our experiments on SpikeToken and analysis of the results against standard tokenizers (BPE, WordPiece, and Unigram) reveal that it is not only a novel spiking implementation of subword tokenization but it also performs strongly on standard tokenizer quality metrics. Particularly, SpikeToken with $k=64$ is a promising choice \emph{among those evaluated}, exhibiting the following summarized characteristics on the 596-document test set:
\begin{enumerate}
    \item It has the lowest token count and thus highest compression level
    \item Its token sequences measure the highest $H_{1}$ and $H_{2.5}$ entropies
    \item It reports the highest vocabulary utilization
    \item It has the highest bytes token density.
    \item From the training timing experiments, increasing $k$ from 1 to 64 reduced training time by a factor of 17 in the current Lacuna implementation.
\end{enumerate}

However, SpikeToken has some limitations. First, the evaluation focused on English--Spanish corpora and did not evaluate additional languages. Future work should explore this, but because SpikeToken is an open-vocabulary subword tokenizer, we do not anticipate any major findings from such studies.

Furthermore, energy measurements were not performed on neuromorphic hardware as access to such hardware is currently limited and the authors do not presently have access to Loihi~2 or SpiNNaker2 chips. This will be an objective of future work.

Most significantly, we highlight a major limitation that is less in SpikeToken itself but in the current state of neuromorphic hardware as it pertains to LLMs. At $V=65{,}536$, our fixed-point Lava circuit uses 67{,}589 neurons and 528{,}895 recurrent synapses. This is within the stated maximum of one million neurons on a single Loihi~2 chip, although physical placement also depends on how the synapses and spike routes are distributed across cores. For SpiNNaker2, the mapper placed vocabularies of up to 4{,}096 IDs on a single chip, with the largest using 132 of the 148 available cores. These results verify tokenization in simulation and provide single-chip placement results for the smaller SpiNNaker2 circuits. However, modern LLMs contain billions or trillions of parameters, each of which maps to a synapse on neuromorphic hardware, exceeding the synaptic capacity of even the largest current neuromorphic systems. This implies that the current state of neuromorphic hardware is not ready to support LLMs. In comparison to the neuromorphic resource needs of LLMs, which are dominated by the synaptic storage of billions or trillions of parameters, the needs of SpikeToken are minimal. Nevertheless, SpikeToken fills an important gap that will enable such objectives when the hardware support becomes available.

% Most significantly, we highlight a major limitation that is less in SpikeToken itself but in the current state of neuromorphic hardware as it pertains to LLMs. Depending on the backend used (Lacuna vs Lava), SpikeToken requires a fairly large number of neurons relative to what current neuromorphic hardware can support. For Lacuna, it ranges from 1{,}558 at $V=512$ to 98{,}395 at $V=32768$ and for Lava fixed-point LIF that same range is 10{,}527 to 1{,}908{,}747. A single Loihi 2 chip can only run one million neurons and thus one or two chips would need to be dedicated to tokenization. By comparison, modern LLMs contain billions or trillions of parameters, each of which maps to a synapse on neuromorphic hardware, exceeding the synaptic capacity of even the largest current neuromorphic systems. This implies that the current state of neuromorphic hardware is not ready to support LLMs.  In comparison to the neuromorphic resource needs of LLMs, which are dominated by the synaptic storage of billions or trillions of parameters, the needs of SpikeToken are minimal. Nevertheless, SpikeToken fills an important gap that will enable such objectives when the hardware support becomes available. 

\section{Conclusion}
In this work, we introduced SpikeToken, a gradient-free spiking tokenizer that learns subword vocabularies and segments bytes into tokens using spike dynamics. During learning, synaptic plasticity accumulates adjacent-pair frequencies and a spike race selects which pairs to merge, with each merge creating the route that SpikeToken later uses to recognize that token in raw bytes. With a single merge per sleep cycle, SpikeToken recovered 65{,}502 of 65{,}536 BPE spellings, reconstructed every one of the 596 held-out documents byte for byte at all vocabulary sizes, and performed comparably to or better than BPE, WordPiece, and Unigram on standard tokenizer quality metrics. We further verified its token decisions in the Lava Loihi~2 and SpiNNaker2 simulators and on a physical Teensy~4.0 microcontroller. Although current neuromorphic hardware cannot yet support practical-scale LLMs, SpikeToken's resource needs scale with vocabulary size rather than model size, positioning it as a building block for future end-to-end neuromorphic language models. All code will be released upon publication acceptance.
% Future work will measure energy and latency on physical neuromorphic hardware, evaluate additional languages and larger corpora, and integrate SpikeToken with a downstream spiking language model.

\bibliography{references}
\end{document}